\documentclass[11pt,a4paper]{erlarticle}

\usepackage{mathtools}
\usepackage{booktabs}
\usepackage{graphicx}
\usepackage{subcaption}
\usepackage[nameinlink,capitalise]{cleveref}

\graphicspath{{figs/}}

\title{Does a Learned Corrector Beat a Simple Retreat?\\
Evidence from a Frozen VLA}
\runningtitle{Runtime-Correction Gains on a Frozen VLA}

\author[1]{Chenchao Sheng}
\author[1]{Zhuang Jiang}
\author[1]{Liuhaichen Yang}
\author[1]{Ningwei Bai}
\author[1]{Zezhi Tang\textsuperscript{*}}
\affil[1]{University College London (UCL), London, United Kingdom}
\titlenote{\textsuperscript{*}Corresponding author: \href{mailto:zezhi.tang@ucl.ac.uk}{\texttt{zezhi.tang@ucl.ac.uk}}}
\date{}

\begin{document}

\maketitle

\begin{abstract}
Before deploying runtime recovery for a frozen vision-language-action (VLA) policy, one must establish that an intervention improves success beyond ordinary run-to-run variation and that its complexity adds value over a simple action. We evaluate these questions on frozen $\pi_{0.5}$ across four RoboTwin tasks. For each test seed, we pair rollouts with and without correction and include a same-seed base-policy re-run as a placebo. Seed-cluster intervals and prespecified comparison rules assess net gains against stochastic outcome changes. Across 3,888 paired episodes, the full pipeline raises success on beat\_\allowbreak block\_\allowbreak hammer by $+13.5$\,pp (95\% interval $[+9.4,+17.7]$), with no detectable gain on the other three tasks at the deployed weight. Among failed base episodes on the responsive task, $43.2\%$ succeed on a plain re-run, compared with $63.5\%$ after correction; many nominal rescues therefore reflect the base policy's own variability. A fixed-time trigger and scripted return to an earlier joint configuration produce a net gain with no detected difference from the learned pipeline across two rounds, although our prespecified equivalence criterion is not met consistently. Pausing and a constant-action control do not yield comparable gains. On this benchmark, the decision to intervene depends strongly on the task, and a paired placebo plus a simple retreat baseline are needed to establish what learned correction contributes.
\end{abstract}


\section{Introduction}\label{sec:intro}

Vision-language-action (VLA) policies are usually deployed frozen: the weights that were trained are the weights that run. When such a policy fails, it often fails quietly. The arm slows, hovers, or repeats a motion without completing the task. A natural remedy is runtime correction: watch the rollout, decide that it is failing, take over for a few dozen steps, and then hand control back. When and how control returns is the \emph{hand-back rule}. Several recent systems report that it raises success rates~\cite{FailSafe,VLACorrector,ReCoVLA}. The base policy, however, is stochastic: in our setting three runs of the \emph{same} seed with no intervention disagree on the outcome for 39--60\% of seeds (Section~\ref{sec:placebo}). A rollout that ``was rescued'' may be one that would have succeeded on a second try anyway.

Three things are missing from how runtime correction is evaluated. First, a placebo. Failure-detection work evaluates alarm accuracy~\cite{SAFE,Sentinel,ActProbe,FAILDetect}, and recovery work reports success before and after intervention. None of the runtime-correction studies we surveyed subtracts how often the base policy's outcome flips on a plain re-run. Second, the simplest counterpart. Learned correctors are compared with the base policy, rarely with a non-learning script under the same trigger and hand-back rule. Retreat-and-retry is well established~\cite{BGR,RewindIL,CycleVLA,SPR,CoRe}, but has not been put head-to-head against a learned corrector under one paired protocol with an equivalence margin. Third, paired statistics. Benchmark practice reports the mean of 100 rollouts per task~\cite{RoboTwin20,SimpleVLARL}. Because the base success rate on identical seeds drifts between rounds (separate launches of the evaluation, Section~\ref{sec:placebo}) by more than binomial sampling explains, unpaired means cannot resolve effects of the size we report (Section~\ref{sec:main}).

\subsection{Related work}
Failure detection scores VLA features with conformal calibration~\cite{SAFE}, action statistics without
training~\cite{Sentinel}, the action space~\cite{ActProbe}, or uncertainty scores without failure
data~\cite{FAILDetect}; our trigger is one task-specific threshold plus five shared constants, judged only by
what follows the alarm. For recovery, PLD is closest in structure to our pipeline: a residual
reinforcement-learning teacher whose recoveries are distilled back into the base~\cite{PLD}. Others fine-tune
a vision-language model (VLM) helper on injected failures~\cite{FailSafe}, re-plan by gradient inside a frozen
$\pi_{0.5}$~\cite{VLACorrector}, train a residual recovery policy from VLM-compiled rewards~\cite{ReCoVLA}, or
augment demonstrations into failure--recovery trajectories~\cite{RACER}. Only two keep the base frozen at
deployment~\cite{VLACorrector,ReCoVLA}, and none of the five reports a same-seed placebo or a non-learning
counterpart under the same trigger and hand-back rule.

The mechanism that reproduces our gain is not new. A scripted retreat can be fired from a value
function~\cite{BGR}, replayed from a VLM-certified checkpoint with a frozen ACT policy~\cite{RewindIL},
reverse-executed to a subtask start with a fine-tuned VLA~\cite{CycleVLA}, or learned by LoRA fine-tuning on
progress stalls~\cite{SPR}. Learned recovery policies handle states outside the training distribution~\cite{BackToManifold} or target safety rather than completion~\cite{RecoveryRL}, and a stateless push-back is a training-free retry~\cite{ProbeAct}. Two comparisons come closest to ours. In CoRe's ablation an anchor rollback, a simulator reset that also restores objects, matches the full
method within about a point on 270 paired episodes, but both conditions are
training-free and the numbers are means without intervals~\cite{CoRe}. In FAR a naive retry trails a learned
retry by 4--6\,pp, but that condition updates weights at test time, with no pairing or
intervals~\cite{FAR}. Harness VLA re-positions the arm with analytic primitives on RoboTwin under a different base policy~\cite{HarnessVLA}. On the data side, interactive imitation learning admits corrections by rule~\cite{RaC,Sirius,IntervenGen,CUPID,Dream2Fix}; ours instead replays each rescue along the rescue branch and two no-rescue branches (Section~\ref{sec:admission}). Distilling a privileged teacher into a sensor-limited student is standard~\cite{LearningbyCheating}, and learned gates for when to ask a human stay within one task with a human fallback~\cite{ThriftyDAgger,AIM,SiriusFleet}.

Adjacent reinforcement-learning control work studies when to update a controller and how to reject disturbances. Tang and colleagues combine actor--critic control with disturbance observers for output stabilization and tracking under mismatched disturbances~\cite{Tang2024Stabilization,Tang2024Tracking}; a time-delay extension adds predictive compensation~\cite{Tang2024Delay}. Bai et al. update an adaptive-dynamic-programming controller only when a state-dependent error crosses a threshold~\cite{Bai2026Event}, and a robotics-focused survey places such methods in the context of manipulator and mobile-robot constraints~\cite{Chan2026ADPSurvey}. Earlier observer-based iterative learning for magnetic bearings and tracking control for roll-to-roll coating address disturbance rejection in physical systems~\cite{Tang2019Bearing,Tang2026Coating}. Closer to learned robot policies, Guided Action Flow uses a critic to guide inference in a frozen VLA, while WAM-OPD distils a world-action model on states visited by its student~\cite{Yang2026GuidedFlow,Yang2026WAMOPD}. These studies motivate the trigger, robustness and distillation aspects of our pipeline; our paired placebo and scripted-retreat comparisons test which parts improve task completion after a brief hand-back.

\subsection{This paper}
\begin{figure*}[!t]
\centering
\includegraphics[width=\textwidth]{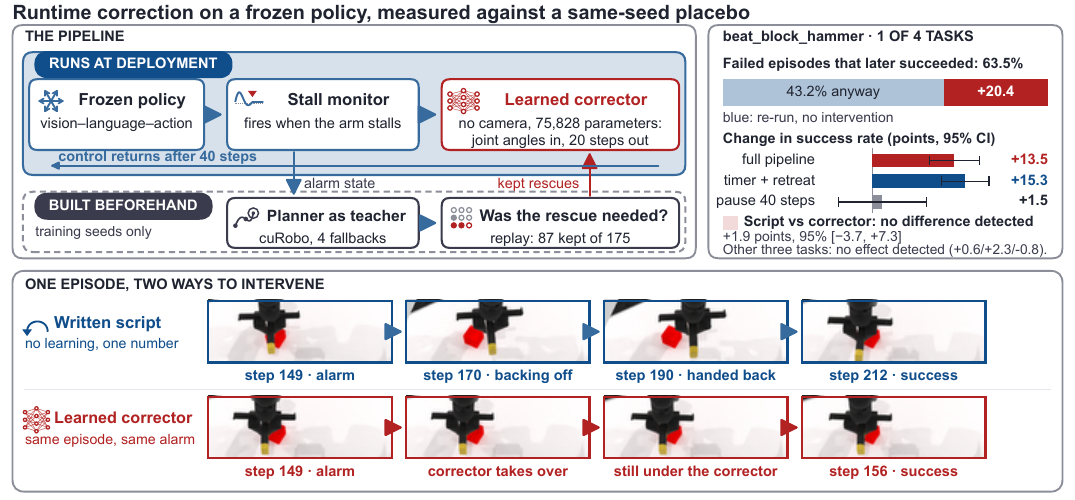}
\caption{The pipeline under test and the protocol that measures it. Left: the frozen policy, the stall monitor and the distilled corrector run at deployment; the privileged teacher and the counterfactual admission step run only on training seeds, where 87 of 175 candidate rescues are kept. Right, bbh: 63.5\% of failed base episodes succeed under the intervention against a placebo of 43.2\%, leaving $+20.4$\,pp; below that, net effects for three conditions with 95\% intervals, then the drift-cancelling difference between retreat and corrector, whose 90\% interval (not drawn) leaves the $\pm 5$\,pp band (Section~\ref{sec:equiv}). Bottom: one bbh test seed, the same alarm at step 149, under the scripted retreat and under the corrector; four frames each, in order and not to scale.}
\label{fig:overview}
\end{figure*}

We fill all three gaps by building a complete pipeline of the kind in~\cite{PLD,FailSafe,VLACorrector,ReCoVLA} and taking it apart. The pipeline (Section~\ref{sec:pipeline}, Fig.~\ref{fig:overview}) keeps $\pi_{0.5}$~\cite{pi05} frozen and adds four components. Two run only during data collection: a privileged teacher that plans rescues with cuRobo~\cite{cuRobo} from the simulator state, and a counterfactual admission step that replays each rescue to check it was needed, keeping 87 of 175. Two run at deployment: a stall monitor, and a camera-free joint-space corrector distilled from the admitted rescues and deployed as one weight file on all four tasks. The protocol (Section~\ref{sec:protocol}) pairs every seed with and without intervention, with a same-seed re-run as the placebo and rules written down before the test seeds ran.

Across 3,888 paired episodes (Section~\ref{sec:main}), the pipeline is significant on one task, beat\_\allowbreak block\_\allowbreak hammer (hereafter bbh, $+13.5$\,pp, 95\% interval $[+9.4, +17.7]$), and indistinguishable from zero on the other three: rotate\_qrcode (qr), place\_a2b\_left (a2b) and place\_a2b\_right (a2br), taken one at a time or pooled (Table~\ref{tab:main}).

Our contributions are three, none of them a new corrector. (i) An evaluation protocol for runtime intervention on frozen policies: per-episode pairing (as in~\cite{CoRe}) and interval reporting (as in~\cite{RaC}), plus a same-seed re-run placebo, a rescue/breakage decomposition with a break-even identity, preregistered equivalence rules and disjoint seed sets. It is needed here: the naive rescue rate is three times the placebo-corrected one. (ii) A component-by-component dissection on the one recoverable task: the gain is reproduced by retreating to the pre-stall configuration. A more elaborate trigger, privileged planning, counterfactual admission and this distilled corrector add no benefit we can detect. (iii) Evidence that, on this benchmark and for the deployed weight, recoverability follows the task, not the repair: a distilled corrector and a hand-written retreat help on the same one task and show no detectable gain on the other three (with exceptions, Section~\ref{sec:tasksagree}). A 324-pair pre-deployment check picks out the one clear case. To our knowledge no prior work makes this comparison on one frozen VLA under a preregistered $\pm 5$\,pp equivalence margin.

\section{Method}\label{sec:pipeline}

\subsection{Trigger: monitor and timer}\label{sec:trigger}
Both triggers share one task-specific quantity, a time threshold $T = \max(100, \lfloor 1.2\,p_{75}\rfloor)$, where $p_{75}$ is the 75th percentile of the step at which successful base rollouts finish on training seeds. This gives 148, 177, 177 and 183 steps on bbh/qr/a2b/a2br. Slash lists follow this order throughout. The \emph{monitor} adds five hand-set constants shared by all tasks (velocity ratio, window, progress fraction, warm-up, earliest snapshot step) and fires when motion inside the window collapses after the threshold. These five were fixed on screening seeds (Section~\ref{sec:tasks}) before any test seed ran. The \emph{timer} keeps only $T$ and fires at step $T$. On 400 held-out seeds with no intervention the monitor missed 0 of 208 failures (95\% exact binomial upper bound 2\%, Clopper--Pearson) and fired on 18\% of successes, ranging from 3\% on a2b to 37\% on a2br. Those false alarms are episodes that would have recovered on their own, which we call \emph{self-healing}, and suppressing them does not pay (Section~\ref{sec:check}).

\subsection{Privileged teacher}\label{sec:teacher}
During data collection only, the teacher attempts a rescue whenever the monitor fires. It is \emph{privileged}: it reads the simulator state, which the deployed corrector never sees. It plans with cuRobo~\cite{cuRobo} through a four-rung fallback sequence: (1) re-plan the scripted grasp-and-place from the current state, (2) retract along a short backtrack and re-plan, (3) retract further and return both arms home before re-planning, and (4) run a task-specific routine for a knocked-over object. The teacher is \emph{not} an upper bound: used online at every alarm it lowers success (Section~\ref{sec:ladder}). Rungs 2 and 3 are themselves retreat-then-retry procedures (Section~\ref{sec:tasksagree}), and they supply 27 of the 87 admitted rescues.

\subsection{Counterfactual admission}\label{sec:admission}
Admission asks whether each rescue was needed and worked. From the alarm state we replay three branches three times each: two no-rescue branches (the base policy continuing) and the rescue branch. A rescue is admitted only if each no-rescue branch succeeds at most once in three and the rescue branch succeeds at least once in three. Because one success in three is enough, the admitted set is permissive: most admitted rescues succeeded once or twice in three, which is one reason the teacher is unreliable when used online. Of 175 teacher rescues on bbh, qr and a2b, 87 pass; the rejected ones are all cases in which the base policy recovered on its own.

\subsection{Distillation, deployment and hand-back}\label{sec:deploy}
The corrector is a 75,828-parameter multilayer perceptron that reads only the 14 joint angles and emits a 20-step joint-angle chunk (280 values). The corrector consists of a LayerNorm layer on the 14-dimensional input, followed by a 256-unit linear layer with GELU activation and a final linear projection to 280 outputs. The network contains 75,828 parameters and requires 0.033 ms per forward pass on CPU. It is trained by full-batch mean-squared-error regression with AdamW~\cite{AdamW} (learning rate $10^{-3}$, weight decay $10^{-2}$) for 600 epochs. When the trigger fires it takes over for two chunks of 20 steps (the \emph{hand-back depth}, 40 steps) and fully replaces the base action (\emph{blend} $1.0$, the weight on the corrector's action), subject to a per-joint step limit. The scripted retreat shares this hand-back rule and differs only in the chunk it emits: a linear interpolation back to the joint configuration of 120 steps earlier, held in a fixed buffer that both triggers share, with no planner and no learning. One weight file, trained on all 116 admitted rescues (the 87 above plus 29 on a2br), is deployed unchanged on every task. This is a deployment constraint, not a claim that multi-task training helps: the ablations show it costs the three original tasks about three points relative to a three-task weight (Table~\ref{tab:ablation}).

\section{Experiments and Results}\label{sec:protocol}

\subsection{Paired design and statistics}\label{sec:paired}
Each test seed fixes the scene, and we run it with the base policy alone and with the intervention. Only discordant pairs (where the two runs disagree) contribute to the net difference. We test them with McNemar's test (the test for paired yes/no outcomes) and report the net effect as the difference in success rate in percentage points (pp). Each seed contributes three such pairs, and McNemar's test treats the repeats as independent, so its $p$-values may be anticonservative. We therefore rely on seed-cluster bootstrap intervals (seeds resampled with replacement, $B = 20{,}000$, repeats never split) and quote $p$-values only as a secondary summary. Where a $p$-value appears with no interval, we report the direction of the effect but do not interpret the $p$-value. We report without multiplicity correction (no adjustment for the number of comparisons). Correction would widen every interval: our null verdicts would stand, our equivalence verdicts would be at risk, and several marginal ablation verdicts (Table~\ref{tab:ablation}) would lose significance, but the bbh effect would not.

\subsection{Placebo and evaluation noise}\label{sec:placebo}
Because the policy samples its actions, three runs of the base policy on the same seed give inconsistent outcomes on 60/51/39/43\% of seeds. The simulator itself is deterministic. With the policy's sampling noise also seeded per episode, three repeats of 100 bbh seeds agree on all 100. Without that seeding they disagree about as often as the main runs do, so the disagreement comes entirely from the policy's sampling. We keep that sampling deliberately: a deployed policy is stochastic, and an intervention changes every state after it. Here $\rho$ is the fraction of failed base episodes the intervention rescues and $\beta$ the fraction of base successes it breaks. The \emph{placebo} rates $\rho_0,\beta_0$ are the same quantities on a same-seed base re-run of the whole episode, and $\rho-\rho_0$, $\beta-\beta_0$ are the placebo-corrected ones.

Across rounds, two runs on identical seeds agree on 56--62\% of episodes, and the rates themselves move: eleven rounds on the same 324 bbh pairs (108 seeds $\times$ 3 repeats) with byte-identical code spread the base success rate over 14.5\,pp. Binomial sampling alone would spread them over about 9\,pp if the 324 pairs were independent, and somewhat more with three repeats per seed, still short of the range observed. We call this spread \emph{drift}. It shows no time trend, and we do not attribute it to a cause.

\subsection{Break-even identity, equivalence rules and preregistration}\label{sec:equiv}
With base success rate $b$, the net effect equals $\rho(1-b) - \beta b$ exactly on any paired table; when $0<b<1$ and $\beta>0$, it is positive if and only if $\rho/\beta > b/(1-b)$. A base re-run has zero net effect by construction ($\rho_0(1-b) = \beta_0 b$). Subtracting the placebo therefore re-splits the net effect between rescue and breakage without changing its size. The identity constrains what we say about where the gain comes from, not how large it is, and as a predictor it is coarse (Section~\ref{sec:check}).

Two conditions are called \emph{equivalent} only if the 90\% interval of their difference lies within $\pm 5$\,pp (the two one-sided tests rule). All other intervals here are 95\%. An interval that crosses zero but leaves the band is reported as ``no difference detected'', never as equivalence. The $\pm 5$\,pp margin was fixed before the eleven-round diagnostic above, from a smaller earlier estimate of the same between-round swing. It is 37\% of the one effect we detect, so ``equivalent'' here means ``not distinguishable at the resolution of this benchmark'', not ``the same''.

The rules, sample sizes and what we would conclude from each outcome went into a ledger before the test seeds ran. An earlier round on 768 development pairs, on a disjoint seed range, pooled $+6.9$\,pp. The main table is a preregistered re-test on test seeds at five times that size, configuration unchanged, where that effect shrank to $+3.9$. Baseline conditions (pause, constant action, monitor + retreat, timer + retreat, planner online), alternative training-run seeds and the robustness conditions reused the main table's seeds but ran as separate launches, so their comparisons are cross-round and read under the 5\,pp band. The instruction-rewording condition ran on the ablation seeds instead. For the cross-round comparisons on which our conclusions rest, we report two statistics: the \emph{endpoint}, the difference in intervention success between rounds, and the \emph{difference of net effects}, which cancels drift shared by the base and intervention runs of a round. Where they disagree we state the weaker conclusion.

\subsection{Tasks, seeds and conditions}\label{sec:tasks}
We selected four RoboTwin 2.0~\cite{RoboTwin20} tasks from its 51-task library; their base success rates are in Table~\ref{tab:main}. The privileged teacher reuses their scripted expert routines. Two of the four selected tasks failed screening criteria that we then chose to override, bbh on the monitor's false-alarm line and a2br on three.

Main table: 972 pairs per task. Ablations: 648 pairs for bbh, qr and a2b against a three-task control. Test seeds are four disjoint blocks of 324 seeds, each giving the main table's 972 pairs with three repeats; the first 108 seeds of a block form its first \emph{shard}, and each block holds three. Training, development, test, ablation, screening and held-out ranges are pairwise disjoint.
\subsection{Main results and the placebo correction}\label{sec:main}
\begin{figure*}[t]
\centering
\includegraphics[width=\textwidth]{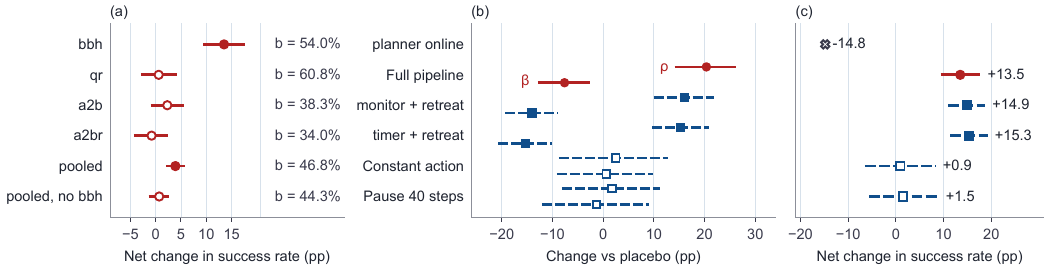}
\caption{The deployed corrector, and the ladder on bbh. Filled markers exclude zero throughout. (a) Net change in success rate per task and pooled, with each task's base rate $b$. (b) Placebo-corrected rescue and breakage. The upper line of each pair, marked $\rho$, is $\rho-\rho_0$; the lower, marked $\beta$, is $\beta-\beta_0$. The two rates have different denominators and do not add. The privileged-planner row is blank here because that condition has no same-seed placebo, and it carries no interval in (c) because its per-episode pairs were not saved. (c) The ladder: six interventions sharing one trigger threshold, with 95\% seed-cluster bootstrap intervals where available. The three retreat and pipeline rows run on 972 pairs, the rest on 324.}
\label{fig:ladder}
\end{figure*}

On 972 pairs per task the pipeline changes success by $+13.5$\,pp on bbh and is indistinguishable from zero on qr, a2b and a2br, alone or pooled (Table~\ref{tab:main}, Fig.~\ref{fig:ladder}a). On bbh the intervention rescued 63.5\% of failed episodes against a placebo of 43.2\%, a placebo-corrected rescue of $+20.4$\,pp. Breakage went the other way: $-7.6$\,pp $[-12.9, -2.5]$ against placebo, that is, fewer base successes were spoiled than a plain re-run spoils. On the other three tasks rescue and placebo rescue are indistinguishable; we call them the \emph{null tasks}: no effect detected at 972 pairs. We have not shown there is none. Every quantity averages over all pairs, including base successes on which no alarm fires. Effects conditional on an alarm are larger, and that dilution makes ``no difference detected'' easier to reach. In break-even terms only bbh's measured $\rho/\beta$ clears its $b/(1-b)$ line with margin (Fig.~\ref{fig:plane}).

\begin{table}[!t]
\centering
\caption{Main table. Four RoboTwin tasks, 972 paired episodes each (324 seeds $\times$ 3 repeats). $b$: base success. Net: intervention minus base success (pp), with a 95\% seed-cluster bootstrap interval and McNemar $p$. $\rho/\rho_0$ and $\beta/\beta_0$: rescue and breakage rates with their same-seed placebo. Placebo-corrected rescue $=\rho-\rho_0$, breakage $=\beta-\beta_0$.}
\label{tab:main}
\footnotesize
\setlength{\tabcolsep}{2pt}
\resizebox{\textwidth}{!}{%
\begin{tabular}{lrrccccl}
\toprule
Task & $n$ & $b$ (\%) & Net (pp) [95\% CI] & $p$ & $\rho/\rho_0$ (\%) & $\beta/\beta_0$ (\%) & Placebo-corr.\ rescue / breakage \\
\midrule
\textbf{bbh} (beat\_block\_hammer) & 972 & 54.0 & \textbf{13.5 [9.4, 17.7]} & $<10^{-4}$ & 63.5 / 43.2 & 29.1 / 36.8 & \textbf{20.4 [14.1, 26.2] / $-$7.6 [$-$12.9, $-$2.5]} \\
qr (rotate\_qrcode) & 972 & 60.8 & 0.6 [$-$2.9, 4.2] & 0.77 & 40.7 / 43.3 & 25.2 / 27.9 & $-$2.6 [$-$8.8, 3.4] / $-$2.7 [$-$7.2, 1.7] \\
a2b (place\_a2b\_left) & 972 & 38.3 & 2.3 [$-$1.0, 5.6] & 0.18 & 22.5 / 20.8 & 30.4 / 33.6 & 1.7 [$-$2.1, 5.6] / $-$3.2 [$-$8.9, 2.2] \\
a2br (place\_a2b\_right) & 972 & 34.0 & $-$0.8 [$-$4.3, 2.5] & 0.68 & 21.5 / 21.5 & 44.2 / 41.8 & 0.0 [$-$4.2, 3.9] / 2.4 [$-$3.7, 8.3] \\
\midrule
\textit{pooled} & 3,888 & 46.8 & 3.9 [2.1, 5.7] & $2.3 \times 10^{-5}$ & 34.4 / 30.0 & 30.9 / 34.2 & --- \\
\textit{pooled without bbh} & 2,916 & 44.3 & 0.7 [$-$1.4, 2.7] & 0.51 & 26.4 / 26.4 & 31.6 / 33.1 & --- \\
\bottomrule
\end{tabular}}
\end{table}

\subsection{Intervention ladder and replication round}\label{sec:ladder}
We ran a ladder of six intervention conditions on bbh under one trigger threshold (Fig.~\ref{fig:ladder}c). Five share the 40-step hand-back rule; the planner condition re-plans to completion instead of handing back. On the first 324 test pairs, where the pipeline gives $+12.7$, pausing the arm and a constant, scene-independent action change success by under two points, neither detectably, and neither rescues at a rate we can distinguish from its placebo. On the endpoint the full pipeline beats each on the same seeds by $+15.7$\,pp $[+8.0, +23.1]$ and $+13.6$\,pp $[+5.6, +21.3]$; on the difference of net effects, $+11.1$ and $+11.7$, both still excluding zero. The scripted retreat under the same monitor changes success by $+17.9$\,pp on those pairs. The privileged planner acting online at every alarm loses about fifteen points with breakage above half; this is the one condition where our own protocol was not met, because its per-episode pairs were not saved, so we read it as a direction only.

On all 972 pairs the three conditions that work are within a few points of each other: full pipeline $+13.5$, monitor + retreat $+14.9$ (the $+17.9$ above was its first shard), timer + retreat $+15.3$, with their placebo-corrected rescue and breakage in Fig.~\ref{fig:ladder}b. On the endpoint, timer + retreat minus the full pipeline is $+0.7$\,pp and monitor + retreat $+1.7$, with 90\% intervals $[-2.6, +4.0]$ and $[-1.4, +4.9]$, both inside the band. On the difference of net effects the same two read $+1.9$ and $+1.4$, and both 90\% intervals leave it, so by the rule of Section~\ref{sec:equiv} we take the weaker verdict of no difference detected.

\begin{figure*}[!t]
\centering
\includegraphics[width=\textwidth]{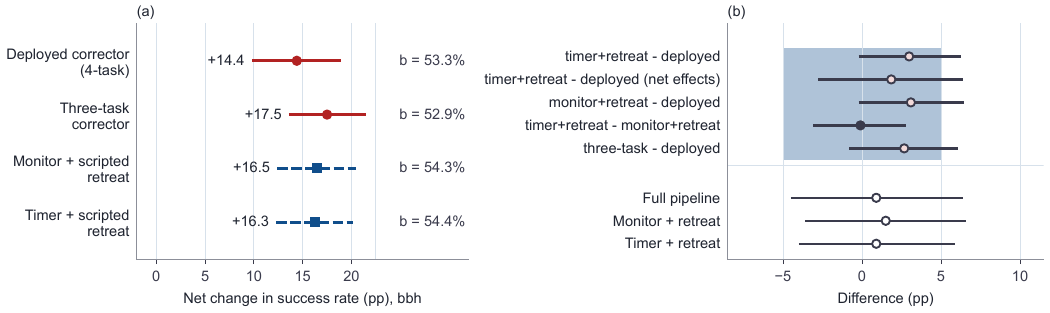}
\caption{Replication round: four conditions on bbh, 972 pairs each, on one GPU in one time window. (a) Net change in success rate, with each arm's base rate $b$; filled markers exclude zero at 95\%. (b) Pairwise differences, five within the round (90\% intervals, upper group) and three between rounds (95\%, lower group), against a shaded $\pm 5$\,pp margin; pale red marks ``no difference detected'', and the one dark marker is the pair of scripted conditions, the only comparison here that is equivalent. Every one of the eight contains zero, and the between-round net effects differ by at most 1.5\,pp.}
\label{fig:replication}
\end{figure*}

Because that comparison is between rounds, we preregistered a replication: four conditions in one round, same code, on all 972 bbh pairs (Fig.~\ref{fig:replication}). All four land within a few points of each other, the two scripted conditions are equivalent, and the three conditions that also ran earlier differ between rounds by at most 1.5\,pp with intervals covering zero. Within the round, timer + retreat minus the deployed corrector gives no difference detected on either statistic, with both point estimates favoring the script. Across two rounds and two statistics, four intervals bear on the deployed corrector against the timer, and only one falls inside the $\pm 5$\,pp band, so we report no difference detected throughout and do not claim equivalence. The point estimates split the net effect differently, the corrector rescuing more and breaking more ($+20.7$ and $-8.9$ against $+18.3$ and $-14.6$), with no interval on that split, and the monitor condition does not follow the pattern. The three-task weight does not change the reading: within the round the three-task corrector sits $+2.7$\,pp $[-1.4, +6.8]$ above the deployed one, an interval that covers both zero and the $7.6$\,pp bbh gap of Table~\ref{tab:ablation}.

\begin{table}[!t]
\centering
\caption{Component ablations. 648 paired episodes per task; control is a three-task corrector trained on the 87 admitted rescues (40-step hand-back). Entries are variant minus control in intervention success, with 95\% intervals. Equivalence was preregistered for the pooled effect only; per-task columns carry no equivalence verdict. Verdict column: \emph{equivalent} = 90\% pooled interval inside $\pm 5$\,pp; \emph{worse} = 95\% pooled interval excludes zero and is negative.}
\label{tab:ablation}
\footnotesize
\setlength{\tabcolsep}{2pt}
\resizebox{\textwidth}{!}{%
\begin{tabular}{lccccl}
\toprule
Ablation & bbh & qr & a2b & pooled & Verdict \\
\midrule
Admission: all 175 vs admitted 87 & $-$1.1 [$-$6.0, 3.9] & 0.5 [$-$3.9, 4.8] & $-$2.5 [$-$6.6, 1.5] & $-$1.0 [$-$3.6, 1.5] & equivalent \\
Admission: random 87 (draw a) vs admitted 87 & $-$4.2 [$-$9.3, 0.9] & $-$0.6 [$-$4.9, 3.7] & $-$2.8 [$-$6.8, 1.2] & $-$2.5 [$-$5.1, 0.1] & equivalent \\
Admission: random 87 (draw b) vs admitted 87 & 0.3 [$-$4.3, 4.9] & $-$0.6 [$-$5.1, 3.9] & $-$5.1 [$-$9.0, $-$1.2] & $-$1.8 [$-$4.3, 0.7] & equivalent \\
Trigger: timer vs monitor & $-$2.0 [$-$6.8, 2.9] & 0.3 [$-$3.9, 4.5] & $-$4.8 [$-$8.5, $-$1.2] & $-$2.2 [$-$4.6, 0.3] & equivalent \\
Teacher: rung 1 only (60 rescues) & $-$4.3 [$-$9.6, 0.9] & $-$2.8 [$-$7.1, 1.5] & 0.0 [$-$4.0, 4.0] & $-$2.4 [$-$5.0, 0.3] & equivalent \\
Hand-back depth: 20 steps vs 40 & $-$5.4 [$-$10.6, $-$0.2] & 4.0 [$-$0.2, 8.2] & $-$3.1 [$-$7.1, 0.8] & $-$1.5 [$-$4.1, 1.1] & equivalent \\
Hand-back depth: 80 steps vs 40 & $-$3.5 [$-$8.6, 1.4] & $-$6.5 [$-$11.1, $-$1.9] & $-$2.5 [$-$6.5, 1.5] & $-$4.2 [$-$6.8, $-$1.5] & worse \\
Training pool: 4 tasks vs 3 tasks & $-$7.6 [$-$12.7, $-$2.5] & $-$2.2 [$-$6.2, 2.0] & 0.0 [$-$4.0, 4.0] & $-$3.2 [$-$5.8, $-$0.6] & worse \\
\bottomrule
\end{tabular}}
\end{table}

\subsection{Component ablations}\label{sec:ablations}
\begin{figure*}[!t]
\centering
\includegraphics[width=\textwidth]{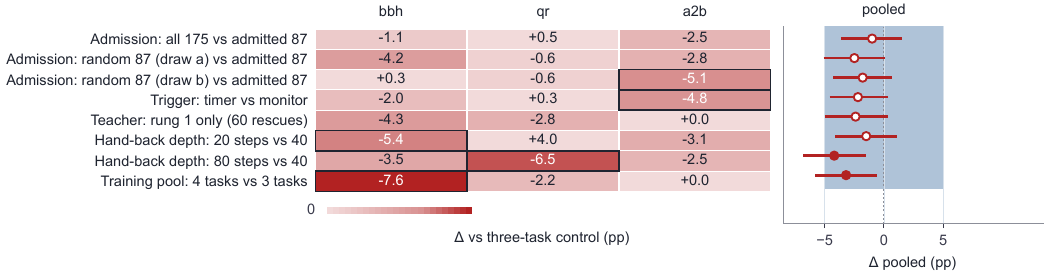}
\caption{Component ablations: each variant minus the three-task control, 648 pairs per task. Left: per task, shaded by how much worse the variant is, so an unshaded cell is an improvement; boxed where the 95\% interval excludes zero, and the intervals themselves are in Table~\ref{tab:ablation}. Right: pooled, filled where the interval excludes zero; bars are 95\% intervals. The shaded band is $\pm 5$\,pp: all eight pooled estimates fall inside it and two are graded worse. Equivalence verdicts use 90\% intervals, which are not drawn (Section~\ref{sec:equiv}).}
\label{fig:ablation}
\end{figure*}

The component ablations (Table~\ref{tab:ablation}, Fig.~\ref{fig:ablation}) reach the same conclusion by removing components rather than adding them, around a different reference point: the three-task control weight rather than the deployed four-task weight, which the fourth task's data weakens (Section~\ref{sec:deploy}). Five variants each cost between one and three points pooled, and all five are equivalence verdicts (Table~\ref{tab:ablation}): removing the admission step, replacing it by either of two independent random draws of 87 from the 175, replacing the monitor by a timer, and keeping only the first teacher rung. Two of them, the timer and the first random draw, have a 90\% interval that nonetheless excludes zero, so we read those as ``no more than a few points, in a consistent direction'', not as ``no effect''.

Per task, where no equivalence verdict applies, the timer and one random draw each cost a2b about five points, both significant. What does change the outcome is hand-back depth, which hurts bbh at 20 steps, marginally, and qr at 80, and the fourth task's data. The scripted retreat moves by about two points across the same range on bbh, with no difference detected between its depths. A corrector trained on bbh data alone comes out below the three-task weight at each of five training-run seeds, significantly for two, one of them marginally. Across three training-run seeds of the deployed weight the bbh effect ranges from $+13.4$ to $+17.4$, and the seed selected on development seeds was not the best on test.

\begin{figure}[t]
\centering
\includegraphics[width=\columnwidth]{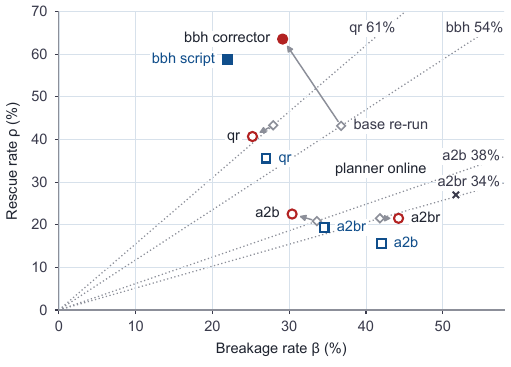}
\caption{Rescue--breakage plane. Circles: the deployed corrector, 972 pairs per task. Squares: the scripted retreat, 972 pairs on bbh (monitor + retreat) and 324 on the others (retreat). Open grey diamonds: the same-seed base re-run, the placebo; arrows run from each placebo to its corrector. The single dark marker is the privileged planner online, bbh only, 324 pairs and no placebo. Dotted: the break-even lines $\rho/\beta = b/(1-b)$, one per task, each labelled with its task and that task's base rate $b$ from the corrector arm. Only bbh sits well above its line under both interventions.}
\label{fig:plane}
\end{figure}

\subsection{Agreement between the two interventions}\label{sec:tasksagree}
If the bbh gain came from the specific corrector, a different intervention might help the others. We detect none: on 324 pairs per task the scripted retreat changes success by $-0.9/-5.6/+1.5$ on qr/a2b/a2br (a2b: $[-11.7, +0.6]$), and no difference from the corrector is detected on any of them (Fig.~\ref{fig:plane}). On a2b the script's breakage exceeds placebo by $12.6$\,pp $[+0.8, +24.2]$ on 324 pairs, marginally: retreating where it does not rescue spoils episodes that would have succeeded. Two preregistered conditions agree. The privileged planner online on a2b changes success by $-2.2$\,pp (counts from the run log, no interval), so it shows no sign of rescuing a2b either. Timer + retreat on the three null tasks gives $-4.6/-0.9/+0.3$ (qr/a2b/a2br), with no detectable difference from the corrector. That completes the grid (corrector, monitor + retreat and timer + retreat on four tasks) in which only bbh is recoverable at the deployed weight and threshold.

Three exceptions qualify this, and none overturns it for the deployed weight, threshold and script. A training-run seed rejected on development seeds gave $+11.4$\,pp on a2br's first shard; a preregistered run on the other two gave $+3.1$ and $+2.8$, neither significant, and the three shards pooled give $+5.8$\,pp ($p = 0.001$). On 648 ablation pairs a2b rises $+4.9$\,pp under the three-task weight and equally under the deployed one (Table~\ref{tab:ablation}, last row), and an earlier trigger ($0.9\times T$) by $+8.3$\,pp on 324 test pairs. All three beat their own base on McNemar alone, but against the deployed configuration only the a2br seed is significant, and only on the difference of net effects, so by Section~\ref{sec:equiv} the weaker verdict stands. The two interventions agree, but rungs 2 and 3 of the teacher are themselves retreats, so the corrector's rescues may be retreat-like, which makes the agreement weaker evidence than two fully independent interventions would give.

\subsection{Pre-deployment check and intervention gate}\label{sec:check}
The check follows from the $\rho/\beta$ plane (Fig.~\ref{fig:plane}): run 324 paired episodes, estimate $b$, $\rho$ and $\beta$, and compare $\rho/\beta$ with $b/(1-b)$. On the first shard it identifies bbh as recoverable ($1.84$ against $1.08$) and classifies qr as not recoverable ($1.31$ against $1.72$). On all 972 pairs, though, qr sits on its line ($1.61$ against $1.55$), so a negative verdict at 324 pairs can reverse by 972. It is inconclusive on a2b and a2br, whose ratios sit on their lines (Fig.~\ref{fig:plane}). Estimating $\rho$ and $\beta$ on the first third of the test seeds and predicting the net effect on the rest gives errors spanning $-7.9$ to $+2.3$\,pp, which exceeds the effects it would need to resolve on a2b and a2br.

Withholding intervention when the alarm is false (an intervention gate) is not a substitute. On 500 base rollouts per task, a classifier separating self-healing alarms from the rest reaches a within-task AUC (area under the ROC curve) of $0.83/0.73/0.52/\approx 0.5$ but only about $0.5$ across tasks (two runs of the same evaluation gave $0.49$ and $0.54$), so it does not transfer. Nor does it pay: a gate on bbh is worthwhile only if more than 68.6\% of the alarms it suppresses were self-healing (its $\rho/(\rho+\beta)$), and the best gate we built reached 63.3\%; sweeping every threshold after the fact tops out at $+0.8$\,pp on bbh. Few alarms self-heal on qr and a2b, so the ceiling is low there, while on bbh a gate this imprecise forfeits more rescues than the breakage it avoids. The self-healing labels come from one rollout per alarm, subject to the same base-policy sampling this paper is about, so these numbers are rough estimates and no better.

\begin{figure}[t]
\centering
\includegraphics[width=\columnwidth]{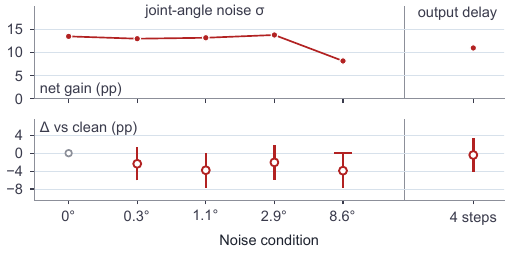}
\caption{Sensitivity of the bbh effect to input noise and output delay, 972 pairs per condition. Top: net gain at four noise levels on the joint angles the corrector reads and at a four-step delay on its output; the leftmost point, at zero noise, is the clean reference. The noise ticks are not to scale. Bottom: the endpoint difference between each condition and the clean one, with 95\% intervals, not the gaps between the points above, which are differences of net effects; the grey marker at zero noise is the clean condition against itself. All five contain zero, and the largest noise level has a bound at exactly zero. Two of the five exclude zero on their 90\% intervals, which are not drawn. Read as differences of net effects, the same five span $+0.3$ to $-5.3$\,pp (Section~\ref{sec:equiv}).}
\label{fig:robust}
\end{figure}

\subsection{Sensitivity to noise, delay, blend and threshold}\label{sec:robust}
The bbh effect survives four levels of noise added to the joint angles the corrector reads and a four-step delay on its output (Fig.~\ref{fig:robust}). Its net effect stays within half a point of the clean condition at the three smaller noise levels, and gives up at most about five points at the four-step delay and at the largest noise. None of the five endpoint differences against the clean run excludes zero at 95\%, so by the rule of Section~\ref{sec:equiv} we report no detectable decay and do not rule out a decay of a few points. Halving the blend of corrector and base action changes the effect by $-4.9$\,pp $[-9.1, -0.8]$ on the endpoint ($-2.6$ as a difference of net effects), the one condition here whose interval excludes zero. Scaling the trigger threshold by $0.9\times$ or $1.5\times$ moves the estimate by two to four points, with no difference detected among the three settings, and rewording the task instruction gives no detectable reduction either. None of this is evidence about a physical robot: the base policy's observations, camera and dynamics were not perturbed.

\section{Discussion}\label{sec:limits}
We compare no external method. The prior systems cited in Section~\ref{sec:intro} differ from ours in trigger, hand-back rule, base policy or benchmark, and none releases a RoboTwin configuration for $\pi_{0.5}$. Our non-learning conditions bound what learning adds under \emph{our} trigger and hand-back rule, not what a different trigger or a vision-conditioned corrector could do. Nor did we run the scene-reset retry we describe below.

One policy, one simulator, four tasks, one of them recoverable; the dissection rests mainly on that task, and no result here was obtained on hardware. On hardware there is no seed, so the placebo becomes a repeat from the same intended scene setup.

Three parts of the deployed configuration were fixed on development seeds: training-run seed, hand-back depth and blend. The script inherited that depth untuned and shows no detectable sensitivity to it, so the ladder compares a tuned corrector against an untuned script. That asymmetry favors the corrector, and the corrector still does not win.

Because $\rho_0$ re-runs the whole episode, it measures what a full retry would achieve: 43.2\% of bbh's failed episodes, more than the $+20.4$\,pp we add on top. A system that could detect end-of-episode failure and reset the scene would reach about $+19.9$\,pp net by the identity of Section~\ref{sec:equiv} rather than by measurement, in the same range as the best conditions we tested. We do not report it as a baseline because our intervention has neither an end-of-episode failure signal nor the ability to reset objects, but a deployment with both should compare against it first.

The intervention takes control of a moving arm without asking, and on a2b the scripted retreat raised breakage 12.6\,pp above placebo (Section~\ref{sec:tasksagree}). On hardware a retreat executed at the wrong moment is a collision risk rather than a lost episode, so the pre-deployment check should be read as a go/no-go safety check and not only as a performance filter. We tested no hardware safety limits.

\section{Conclusion}\label{sec:conclusion}

Our pipeline helps on one of four RoboTwin simulation tasks. There, about two-thirds of the naive rescue rate is what the base policy would have done anyway. A one-parameter timer with a 40-step scripted retreat reaches a net effect we cannot distinguish from the whole pipeline's, in two rounds and under two statistics. Pausing, a constant action, and the privileged planner acting online do not. On the other three tasks neither intervention is distinguishable from zero at the deployed weight, and on one of them retreating may raise breakage above placebo. The practical reading: before deploying any runtime correction, run a few hundred paired episodes with a same-scene placebo and check whether the task clears its break-even line. If it clears with margin, returning the arm to where it was before it stalled may already be enough. We will release the protocol, ledger, launch scripts and result files.

\bibliographystyle{plainnat}
\bibliography{references}

\end{document}